\documentclass[conference]{IEEEtran}
\IEEEoverridecommandlockouts
\usepackage{cite}
\usepackage{amsmath,amssymb,amsfonts}
\usepackage{algorithmic}
\usepackage{textcomp}
\usepackage{subfigure}
\usepackage{float}
\usepackage{graphicx}
\usepackage{color}
\usepackage{xcolor}

\usepackage{booktabs}
\def\BibTeX{{\rm B\kern-.05em{\sc i\kern-.025em b}\kern-.08em
    T\kern-.1667em\lower.7ex\hbox{E}\kern-.125emX}}
\begin{document}

\title{A Multi-Agent Probabilistic Inference Framework Inspired by Kairanban-Style CoT System with IdoBata Conversation for Debiasing}

\author{
\IEEEauthorblockN{Takato Ueno}
\IEEEauthorblockA{
Graduate School of Data Science \\
Shiga University \\
Hikone, Japan \\
s7025101@st.shiga-u.ac.jp
}
\and
\IEEEauthorblockN{Keito Inoshita}
\IEEEauthorblockA{
Faculty of Business and Commerce \\
Kansai University, Osaka, Japan \\
Data Science and AI Innovation \\
Research Promotion Center \\
Shiga University, Hikone, Japan \\
inosita.2865@gmail.com
}
}
\maketitle

\begin{abstract}
Japan's kairanban culture and idobata conversations have long functioned as traditional communication practices that foster nuanced dialogue among community members and contribute to the formation of social balance. Inspired by these information exchange processes, this study proposes a multi-agent inference framework (KCS+IBC) that integrates multiple large language models (LLMs) to achieve bias mitigation, improved explainability, and probabilistic prediction in sentiment analysis. In addition to sequentially sharing prediction results, the proposed method incorporates a mid-phase casual dialogue session to blend formal inference with individual perspectives and introduces probabilistic sentiment prediction. Experimental results show that KCS achieves accuracy comparable to that of a single LLM across datasets, while KCS+IBC exhibits a consistent decrease in entropy and a gradual increase in variance during the latter stages of inference, suggesting the framework's ability to balance aggregation and diversity of predictions. Future work will quantitatively assess the impact of these characteristics on bias correction and aim to develop more advanced sentiment analysis systems.
\end{abstract}


\section{Introduction}
Research in natural language processing (NLP) supports dialogue systems, document summarization, sentiment analysis and machine translation and it finds rapid real-world adoption across society [1]. Recent advances in large language models (LLMs) let us interpret ambiguous expressions and infer based on context, tasks that conventional methods could not handle, and they improve accuracy and flexibility in language understanding [2]. These benefits now reach all sectors. Companies use them for internal document creation and marketing efforts. Educators apply them to personalized learning plans and mental health support around the world [3, 4].Researchers have long noted that language model outputs embed bias from their training data and conceal reasoning steps, which raises serious reliability and fairness concerns [5]. Such issues pose unacceptable risks in fields that demand justifiable decisions, for example policy making, medical diagnosis and recommendation systems, and we must address them without delay. Studies report bias in document classification and opinion analysis that arises from dataset imbalance and the learning process, and developers have proposed numerous debias methods [6, 7].

Despite many proposals, no existing method suffices to debias LLM outputs fully. Sentiment analysis faces especially high risk because it depends on ambiguous expressions and subjective context. Relying on a single LLM can drop diverse viewpoints and eliminate minority opinions and it can undermine the final judgment's reliability and explainability [8]. We need a new multi agent inference framework that unites diverse perspectives and raises explainability while ensuring reliable reasoning. Yet researchers have paid little attention to this approach.

In this study, we propose a multi-agent probabilistic inference framework inspired by a kairanban-style CoT system with idobata conversation to address the aforementioned issues. This framework is based on two traditional Japanese communication cultures: the Kairanban-style CoT System (KCS), which emulates the circular bulletin board practice where information is sequentially passed among members who each add or revise content to build consensus, and the Idobata Conversation (IBC), which draws from informal, free-flowing group chats that allow for the emergence of tacit knowledge through diverse perspectives.Specifically, in KCS, agents sequentially share their opinions using a chain-of-thought (CoT)[9] approach starting from the first agent. Each subsequent agent supplements or corrects the previous output, which helps reduce individual biases and enhance reliability. The use of logs also improves the explainability of the opinion formation process.In IBC, agents engage in free conversational exchanges, enabling the extraction of minority opinions and subtle nuances that are often missed by formal CoT procedures. This contributes to the diversity and flexibility of the overall opinion formation process.

The outputs from these two components are integrated into a probabilistic inference framework that reflects each agent's contribution as a distribution, thereby enabling the fusion of formal reasoning and subjective viewpoints. This framework achieves reliable inference based on probabilistic distributions.In this study, we adopt sentiment analysis as a case study, as it involves a high degree of ambiguity and subjectivity and is easily influenced by bias. We quantitatively verify the effectiveness of the proposed approach in this context.Our main contributions are as follows.

\begin{enumerate}
    \item We integrate probabilistic inference into a culture-inspired multi-agent framework to visualize inter-agent reliability on a task and to demonstrate an effective path to bias reduction.
    \item We introduce KCS, in which agents augment and correct outputs sequentially to aggregate views, reduce bias and improve explainability while forming a reliable consensus.
    \item We introduce IBC, in which agents exchange free-form opinions to capture minority views and subtle nuances and to add diversity and flexibility to the opinion formation process.
\end{enumerate}

The remainder of this paper follows. Section II reviews bias in language models and research on chain of thought. Section III details our proposed multi agent probabilistic inference framework. Section IV describes the experimental setup and datasets. Section V presents quantitative and qualitative evaluation results and discussion. Section VI concludes with our findings and future directions.

\section{Related Works}
\subsection{Methods for Bias Mitigation in LLMs}
In recent years, many methods have been proposed to evaluate and mitigate social biases embedded in LLMs. Gallegos et al. [10] formulated the concepts of social bias and fairness and presented three taxonomies that comprehensively classify evaluation metrics, datasets, and mitigation techniques. Similarly, Ranjan et al. [11] reviewed bias in terms of intrinsic and extrinsic factors, organized the advantages and limitations of existing mitigation techniques, and outlined future research challenges.

Specific mitigation methods include the work of Dixon et al. [12], who proposed a bias reduction approach focused on input-level preprocessing through data correction and filtering. Dwivedi et al. [13] applied prompt engineering tailored to bias mitigation targets, and White et al. [14] systematized effective patterns in prompt engineering. Guo et al. [15] implemented an approach for automating the generation of bias-mitigating prompts, while Ye et al. [16] proposed a method to automatically refine prompts using meta-prompts. Other strategies include adversarial training during model learning, as proposed by Zhang et al. [17], and post-processing techniques to modify outputs after generation, as seen in the work of Yifei et al. [18]. In addition, Felkner et al. [19] developed community-driven datasets for bias evaluation, facilitating a framework that supports both assessment and improvement.

These studies mainly target bias detection and mitigation within a "single model" and have not sufficiently explored bias suppression through the integration of judgments from multiple perspectives. To address this gap, the present study investigates the potential of bias mitigation through multi-agent collaborative inference.
\subsection{Inference Using CoT and Conversational Dialogues}
CoT prompting significantly improves complex reasoning performance in LLMs by encouraging the generation of intermediate reasoning steps. Wei et al. demonstrated that providing CoT-augmented few-shot prompts in mathematical and commonsense reasoning tasks dramatically enhances the performance of models with tens to hundreds of billions of parameters. Building on this, Zhang et al. [20] proposed a framework called "Auto-CoT," which automatically generates and selects diverse CoT demonstrations using the LLM itself, achieving performance on par with human-crafted prompts while eliminating manual effort. To analyze the effectiveness of CoT, Wang et al. [21] identified elements such as "relevance" and "consistency" as key contributing factors. Jin et al. [22] introduced a method that integrates external knowledge inference into CoT using knowledge graphs, improving both knowledge integration and reasoning accuracy.

Meanwhile, research incorporating conversational-style dialogues has also progressed. Chae et al. [23] proposed a method called Dialogue CoT Distillation, which extracts shared understanding from multiple utterances and distills it into dialogue-specific CoT. Wang et al. [24] introduced Cue-CoT, a method that infers user's latent intentions from dialogue context to improve the "usefulness" and "acceptability" of responses. Chen et al. [25] proposed ChatCoT, a framework that enhances CoT reasoning with tool-augmented dialogue turns, where each turn alternates between tool operations and logical thinking. Chang and Chen [26] presented SalesAgent, a CoT agent specialized in sales dialogues. Using the SalesBot 2.0 dataset, SalesAgent improves dialogue coherence and controllability by detecting user intentions and selecting strategies through CoT reasoning.

These attempts to combine CoT and conversational dialogues offer valuable insights into enabling LLMs to perform layered and context-sensitive reasoning that cannot be captured by single-pass inference alone. Building on this, our study proposes a novel framework that integrates Japan's cultural practices—the Kairanban-style CoT and Idobata session—to realize collaborative and diversified judgment structures in sentiment analysis tasks, simultaneously achieving bias reduction and improved explainability.

\begin{figure*}[t]
  \centering
  \includegraphics[scale=0.4]{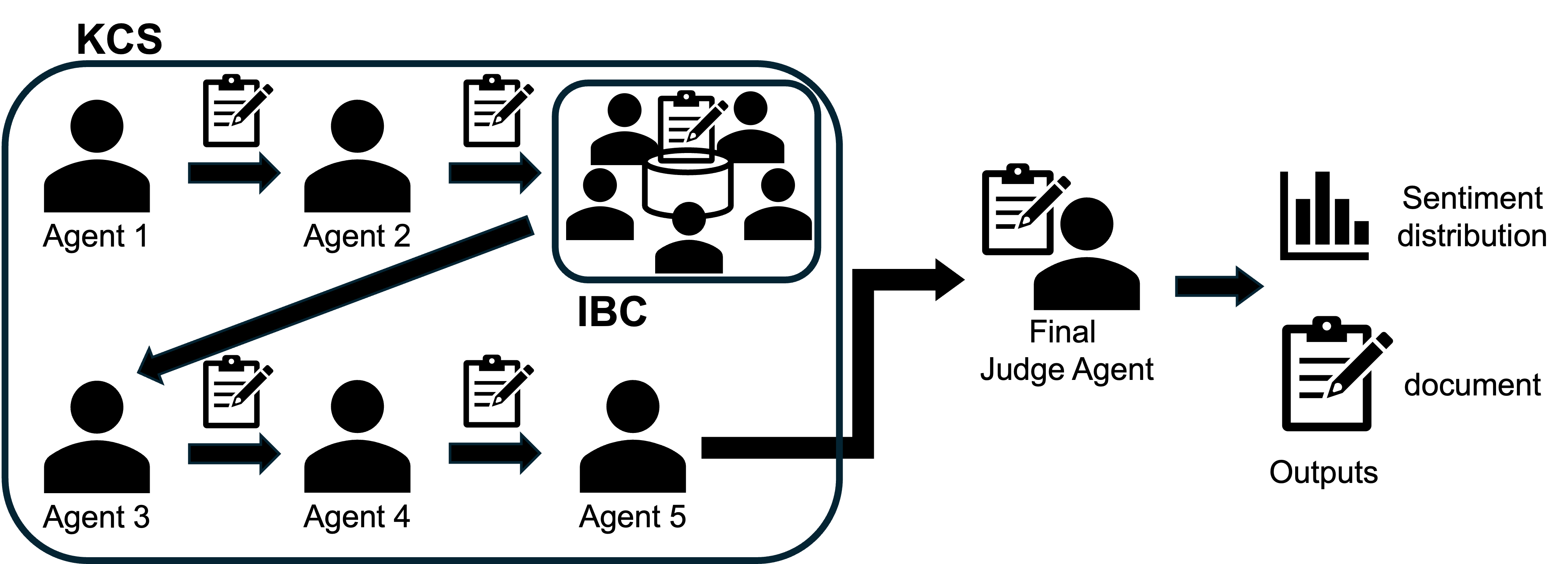}
  \caption{Framework Overview}
\end{figure*}

\section{A Multi-Agent Probabilistic Inference Framework Inspired by Kairanban-Style CoT System with IdoBata Conversation}

\subsection{Framework Overview}
In this study, we introduce a multi-agent probabilistic inference framework inspired by the kairanban-style CoT system with idobata conversation into LLMs, in order to address the limitations of single models, such as insufficient confidence and lack of diverse perspectives. The framework overview is shown in Fig. 1.

First, an initial agent performs a preliminary inference on the input text. Then, through a sequential circulation process, the inference results based on CoT are shared among multiple agents in what we call the KCS. At each step, the current agent outputs its own inference results along with a probability distribution, using the previous agent's output as input.

In the middle of this sequential inference process, we insert a series of informal dialogue sessions, referred to as IBC, which are distinct from the formal CoT descriptions. During IBC, agents freely exchange casual comments with one another. This allows for the extraction of minority opinions and subtle nuances, enriching the opinion formation process with diversity and flexibility.

Finally, the initial agent integrates the most recent CoT outputs produced after the free-form dialogue and makes the final judgment. This enables the realization of multilayered consensus formation that balances high confidence with diverse viewpoints.

\subsection{KCS Based on CoT Inference}

To enhance the diversity and depth of CoT-based inference that accumulates through interactive dialogue, we propose the KCS. This mechanism allows each agent to generate enriched CoT by layering new perspectives upon the analysis results and hypotheses from previous agents. The architecture of KCS is illustrated in the left side of Fig. 1.

Let the agents be indexed by $i = 0, ..., N$, where $N$ is the total number of agents. The initial agent $A_0$ generates a document $D_0 = \{ R_0, S_0, P_0 \}$, consisting of an initial analysis result $R_0$ and a hypothesis $S_0$. Here, $S_0$ is a placeholder used to maintain the prompt structure as an "initial hypothesis" and is intentionally fixed so as not to influence the actual inference process. Similarly, all probabilities in $P_0$ are set to zero as a placeholder.

From step $i \geq 1$, each agent $A_i$ receives the document $D_{i-1}$ output by the previous agent and updates it by obtaining its own analysis result $R_i$, logical reasoning $S_i$, and probability distribution $P_i$ according to the following expression:
$$
(R_i, S_i, P_i) = \textrm{Agent}_i(R_{i-1}, S_{i-1}, P_{i-1})
$$
Through this process, each agent is able not only to generate a single optimal solution but also to maintain multiple possible inferences.

The analysis results, logical reasoning, and probability distributions accumulated in this stepwise manner enable multifaceted and time-sequenced reasoning that is difficult to achieve through one-shot inference. In this framework, each probability distribution obtained at each step is carried over as input to the next step, allowing information on past uncertainty and candidate diversity to be retained. This not only enables parallel consideration of different perspectives and hypotheses but also supports a probabilistic grounding of why a particular conclusion can be considered valid.

Specifically, the initial hypothesis is gradually verified and re-evaluated as it passes through each agent's reasoning, and ultimately, the agents can select the optimal solution while comparing multiple candidates. At the final step $i = N$, the final agent also generates a probability distribution $P_N$, making it possible to explicitly quantify the degree of confidence in the final candidate. For instance, when one candidate has a higher probability than others, but alternative viewpoints also retain some likelihood, this uncertainty can prompt further discussion or additional validation. As a result, the entire reasoning process leading to the final decision becomes transparent, allowing researchers or decision-makers to trace how confidence evolved at each step and explain clearly why a given conclusion was reached.

Thus, KCS does more than merely present output results. It enhances the reliability and persuasiveness of reasoning, and its greatest advantage lies in its robustness against complex problem settings and high-uncertainty situations.

Figure 3 shows the input prompt used during KCS and the expected output format. In this prompt, the result of sentiment analysis by the previous agent is presented first, followed by the opinions of all prior agents in sequence. The new agent is then instructed to express a one-sentence CoT explaining "why they selected that sentiment," to compare it with the previous agent's view, and finally to output the probability distribution of the sentiment labels they computed.

This prompt design ensures that each agent clearly presents their own reasoning while taking into account prior uncertainty. They also highlight differences from the previous stage and points of re-evaluation, and they share the confidence of their judgment as probability values. In doing so, reproducibility and transparency in the reasoning process are greatly improved, maximizing the advantages of constructing a dialogical, multi-stage CoT framework.

\subsection{IBC Based on informal dialogue}

By introducing IBC as an informal dialogue phase in the middle of the sequential inference process guided by KCS, we aim to extract minority opinions and subtle nuances that are often difficult to capture through formal CoT-based reasoning. This enhances the diversity and flexibility of the decision-making process. The architecture of IBC is illustrated in the center of Fig. 1.

During the execution of the KCS module, an IBC session begins at the point when the agent index $i = m$ is reached. From that point on, agents $j = 0, 1, \dots, N$ sequentially generate comments $C^j$. Each agent $j$'s comment is generated based on the following inputs:

\[
C^j =
\begin{cases}
  \mathrm{Agent}_{j}\bigl(x,\;R^{j}\bigr),
    & j = 0,\\[6pt]
  \mathrm{Agent}_{j}\bigl(x,\;R^{j},\;R^{j-1},\;\bigcup_{k\le j}C_k\bigr),
    & 0 < j < m,\\[6pt]
  \mathrm{Agent}_{j}\bigl(x,\;R^{j-1},\;\bigcup_{k\le j}C_k\bigr),
    & j = m,\\[6pt]
  \mathrm{Agent}_{j}\bigl(x,\;\bigcup_{k\le j}C_k\bigr),
    & j > m.
\end{cases}
\]
After the session concludes, all informal comments $\bigcup_{k\le N}C_k$ are appended to the document:
\[
D_{m} = D_{m} \cup \left\{\bigcup_{k\le N}C_k\right\}
\]
\noindent Then, the process returns to the CoT sharing phase in KCS, and based on the updated CoT, transitions to the final integration phase.

By incorporating IBC sessions into KCS, we integrate structured CoT sharing with informal conversational interactions, thereby establishing a multi-layered consensus-building process that balances transparency and persuasiveness while mitigating bias. IBC sessions extract implicit knowledge and subtle nuances that formal CoT-based reasoning cannot capture, allowing agents (with $j < m$) to mutually adjust their probability distributions $P_i$.

Specifically, at the start of the session, agents prior to $m$ already possess multiple candidates with associated probabilities. Through open-ended conversation, they candidly discuss why certain candidates were supported or overlooked. In this interaction, a candidate considered low-probability by one agent might be reevaluated due to insights provided by others. For instance, contextual sarcasm or social implications behind a minority sentiment—elements that are easy to overlook—can be highlighted by other agents, leading to renewed support for a previously dismissed candidate. This process demonstrates that probabilistic inference is not merely a numerical summation, but a refinement process through communication.

\begin{figure}[h]
  \centering
  \includegraphics[width=1\columnwidth]{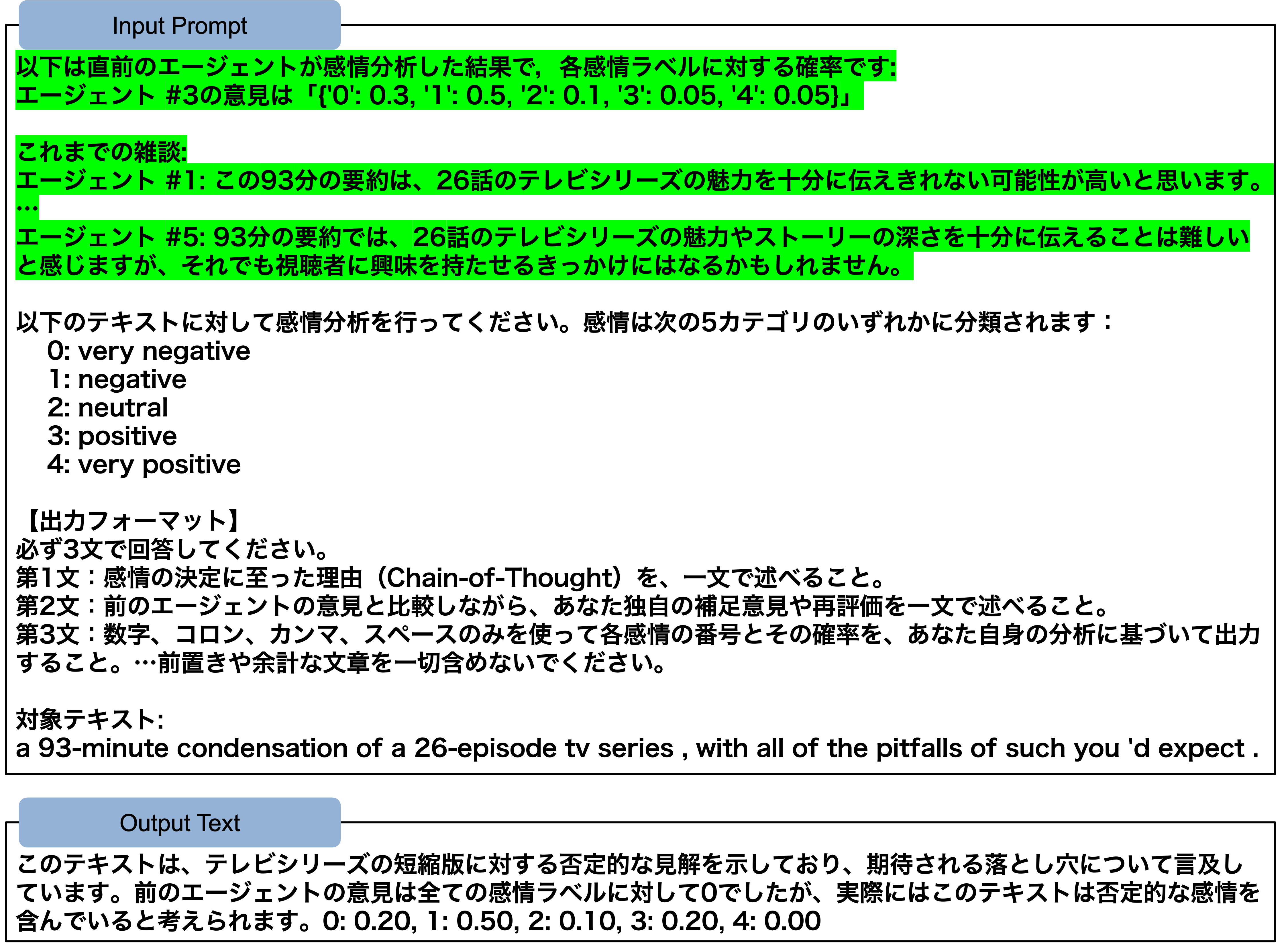} 
  \caption{Input Prompt and Output Text for KCS}
\end{figure}
\begin{figure}[h]
  \centering
  \includegraphics[width=1\columnwidth]{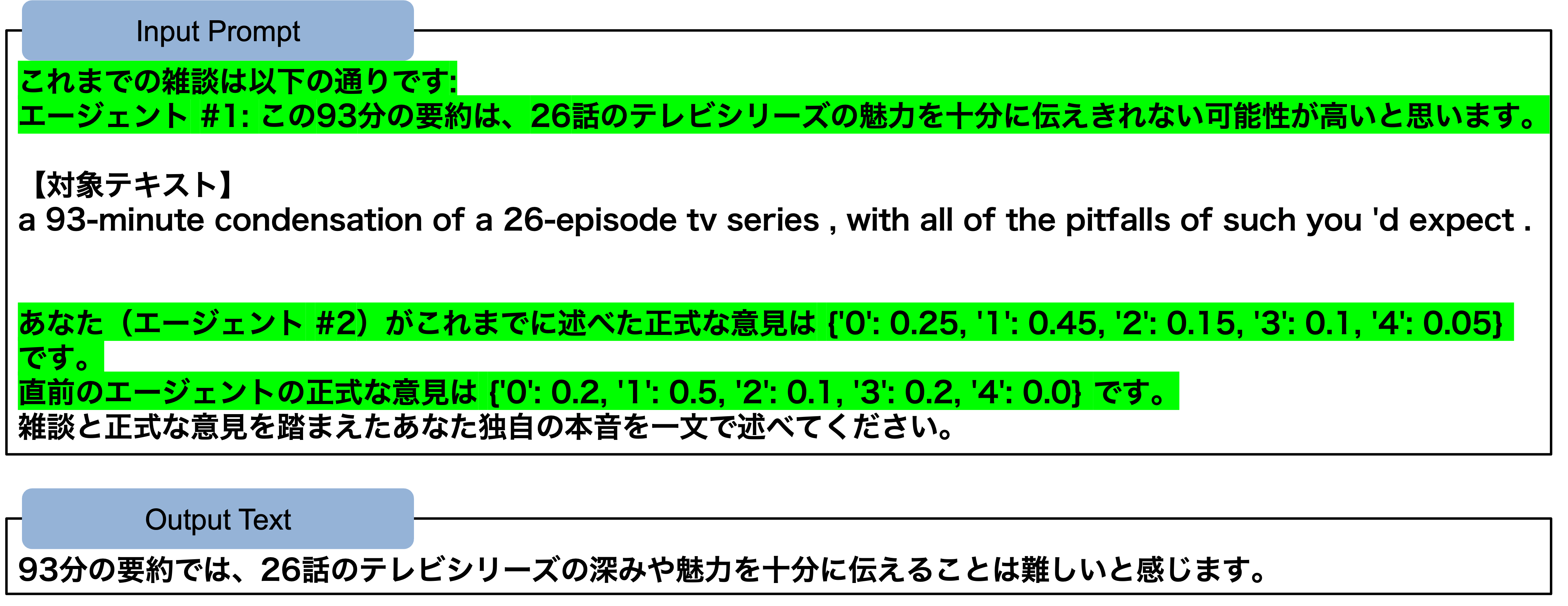} 
  \caption{Input Prompt and Output Text for IBC}
\end{figure}

The advantage of the multi-agent structure lies in its ability to diffuse individual biases and perspective imbalances. In IBC sessions, agents proactively share perspectives they may have previously missed, drawing out minority opinions and personal viewpoints that would otherwise remain buried in formal CoT. Consequently, the final output does not reflect a single agent's most frequent judgment, but rather a synthesis of diverse perspectives from multiple agents. The model thus achieves a balanced judgment that enhances convergence without sacrificing diversity. IBC sessions supplement the formal CoT sharing phase by introducing alternative viewpoints, significantly improving the reliability and explainability of the final decision.

The input prompts used during IBC, as well as example outputs, are shown in Fig. 3. These prompts are designed to encourage each agent to refer to past CoT outputs and freely express opinions during the mid-stage informal dialogue. Agents are not constrained by previous logical structures, and instead, candidly present subtle nuances and minority perspectives. As a result, informal viewpoints are incorporated into the inference process, securing both diversity and transparency.

\subsection{Application to Sentiment Analysis Tasks}

In this study, we apply the proposed method to sentiment analysis, a representative task in natural language processing, to evaluate its performance and characteristics. In sentiment analysis, the primary objective is to estimate the conditional probability distribution $p(Y|x)$ over a set of sentiment labels $Y = \{y_1, \dots, y_K\}$ given an input text $x \in X$, where $K$ denotes the number of sentiment classes. In addition to this probabilistic prediction, our approach also aims to ensure transparency in reasoning by retrieving CoT descriptions generated during the inference process.

The sentiment analysis process under the proposed framework consists of the following three stages:
\begin{enumerate}
    \item KCS Phase: The initial agent outputs a document $D_0 = \{R_0, S_0, P_0\}$, which is then passed on to the next agent. At each subsequent step $i \ge 1$, agent $i$ refines and supplements the previous document $D_{i-1}$, updating it to $D_i$. This stage focuses on hypothesis elaboration and deepening of justification.
    \item IBC Phase: When the sequential reasoning by multiple agents reaches a midpoint (at an arbitrary step $i = m$), the KCS process is temporarily paused. At this point, a dialogue phase is introduced, during which all agents engage in free-form, informal commentary. Each agent generates observations, reactions, or supplementary opinions based on prior sentiment predictions and interpretations offered by others. This session aims to uncover implicit perspectives on sentiments that may have been overlooked, resulting in comments $C^j$ from each agent $j$. These comments are then integrated into the overall document.
    \item Post-IBC KCS Phase: After the IBC session concludes, the process returns to the KCS phase, where subsequent agents perform further reasoning based on the updated document. Finally, using the aggregated document $D_N$ compiled from all agents' outputs, a final judge agent produces the conditional probability distribution $p(Y|x)$ over sentiment labels. Through this process, the model achieves multi-layered and explainable reasoning that incorporates viewpoints and ambiguities difficult to capture with a single LLM.
\end{enumerate}

In this way, KCS functions as the foundation for sequential formal reasoning, while IBC serves as an extension mechanism to introduce diverse perspectives. By integrating both, the proposed multi-agent sentiment analysis process simultaneously controls confidence and diversity in a coherent manner.

\section{Experimental Setup}
\subsection{Objective and Comparison Systems}
The primary objective of this study is to compare and evaluate the accuracy and bias-mitigation effectiveness of the probability distributions over sentiment labels generated by a multi-agent probabilistic inference framework using multiple LLMs for sentiment analysis tasks. In particular, we aim to clarify how the proposed method, which incorporates the sequential and collaborative decision-making structure inspired by Japanese community bulletin board culture and neighborhood conversations, improves the stability and explainability of outputs compared to conventional single LLM-based methods.

This experiment compares the following three systems:

\begin{enumerate}
    \item System A (single): This method performs a single-pass sentiment analysis on the input text and directly outputs a probability distribution. Since it reflects the pure performance of the LLM, it serves as the most basic baseline for performance comparison.
    \item System B (KCS): This system adopts a structure in which multiple agents sequentially append CoT reasoning and probability distributions of sentiment labels, starting from an initial agent and passing them to the next. This enables performance comparison that isolates the effect of the KCS mechanism without incorporating IBC.
    \item KCS+IBC (Ours): In addition to the structure of System B, this system inserts an IBC session mid-process to promote the sharing and re-evaluation of diverse perspectives. By comparing with the above two systems, we aim to demonstrate the individual and combined effects of KCS and IBC.
\end{enumerate}

Through the comparison of these three systems, we investigate how sequential information transmission and the introduction of informal sessions influence output diversity and confidence.

\subsection{Datasets, Input Conditions, and Evaluation Metrics}

We used three benchmark datasets commonly employed in sentiment analysis: the Stanford Sentiment Treebank (SST5) [27], Tweet Eval [28], and the Financial Phrase Bank [29]. From each dataset, we randomly sampled 500 instances to form the test set, balancing computational cost and comparability. Each text in the datasets has a single ground-truth label, which we treated as a one-hot vector to serve as a reference. We then evaluated how each system's output probability distribution over sentiment labels deviated from this reference from multiple perspectives.

All systems were executed using the same input texts. To ensure reproducibility, we fixed the temperature parameter at 0.0 for all model executions. For classification performance evaluation, we used both macro-averaged and micro-averaged F1 scores to analyze the alignment between predicted and true labels in more detail. This approach allows us to account for the effects of class imbalance and biases toward majority classes.

Furthermore, to assess the reliability and calibration of the output probabilities, we adopted cross-entropy loss (log loss) and the Brier score. These metrics evaluate the appropriateness of the probability values output by each system.

Additionally, we computed the mean entropy and variance of the probability distributions at each step to quantitatively analyze how the introduction of KCS and IBC affects the concentration and variability of the distributions. This analysis reveals dynamic characteristics unique to collaborative inference that cannot be captured by classification accuracy alone.
\section{Experimental Results}
\subsection{Classification Performance}
In this study, we compared the performance of the single model and the proposed method using three datasets.

In both the Financial Phrase Bank (Table 1) and Tweet Eval (Table 3), the single model achieved the highest performance in terms of both macro-F1 and micro-F1 scores. On the other hand, in SST5 (Table 2), a more challenging five-class classification task, the proposed method KCS outperformed the single model. Specifically, KCS achieved a macro-F1 score of 0.4488, exceeding the single model's 0.3704 ($\Delta = +0.0784$), and a micro-F1 score of 0.4668, surpassing the single's 0.4425 ($\Delta  = +0.0243$). In addition, LogLoss improved from 7.7227 to 4.2133, and the Brier score improved from 0.1553 to 0.1463. These results indicate that KCS is effective not only in terms of classification performance but also in the calibration accuracy of probabilistic predictions.

Furthermore, KCS+IBC also demonstrated superior performance to the single model in SST5 (Table 2), recording a macro-F1 score of 0.4044 and a micro-F1 score of 0.4668. However, compared to KCS, the macro-F1 score showed a slight decrease. This may be attributed to the introduction of a conversational session in the intermediate phase, which incorporated minority opinions and diverse perspectives into the output, leading to a more averaged and dispersed decision. In other words, KCS+IBC realizes a framework that respects diversity and fosters collaborative reasoning, while maintaining a competitive level of classification accuracy.

\begin{table}[H]
  \centering
  \setlength{\tabcolsep}{4pt}
  \small
  \caption{Financial Phrasebank}
  \label{tab:key_metrics_financial_phrasebank}
  \begin{tabular}{lrrrr}
    \toprule
    model    & macro-F1 & micro-F1 & logloss & brier  \\
    \midrule
    single   & 0.9222   & 0.9297   & 0.4830  & 0.0459 \\
    KCS      & 0.9110   & 0.9177   & 0.4292  & 0.0501 \\
    KCS+IBC  & 0.8002   & 0.8032   & 0.6268  & 0.1004 \\
    \bottomrule
  \end{tabular}
  \vspace{1em}
  \centering
  \setlength{\tabcolsep}{4pt}
  \small
  \caption{SST5}
  \label{tab:key_metrics_sst5}
  \begin{tabular}{lrrrr}
    \toprule
    model    & macro-F1 & micro-F1 & logloss & brier  \\
    \midrule
    single   & 0.3704   & 0.4425   & 7.7227  & 0.1553 \\
    KCS      & 0.4488   & 0.4668   & 4.2133  & 0.1463 \\
    KCS+IBC  & 0.4044   & 0.4668   & 5.3116  & 0.1483 \\
    \bottomrule
  \end{tabular}
  \vspace{1em}
  \centering
  \setlength{\tabcolsep}{4pt}
  \small
  \caption{Tweet Eval}
  \label{tab:key_metrics_tweeteval}
  \begin{tabular}{lrrrr}
    \toprule
    model    & macro-F1 & micro-F1 & logloss & brier  \\
    \midrule
    single   & 0.6450   & 0.6536   & 1.3272  & 0.1731 \\
    KCS      & 0.6338   & 0.6454   & 1.0381  & 0.1790 \\
    KCS+IBC  & 0.5906   & 0.6082   & 1.1472  & 0.1923 \\
    \bottomrule
  \end{tabular}
\end{table}

\subsection{Dynamic changes in entropy and variance}
\begin{table}[b]
  \centering
  \label{tab:kcs_step_entropy_variance}
  \setlength{\tabcolsep}{4pt}
  \small

  \caption{Changes in entropy}
  \begin{tabular}{crr}
    \toprule
    Agent & KCS & KCS+IBC \\
    \midrule
    1 & 0.8377(+0.0000) & 0.8229(+0.0000) \\
    2 & 0.8564(+0.0187) & 0.8336(+0.0107) \\
    3 & 0.8413(-0.0151) & 0.8159(-0.0177) \\
    4 & 0.8488(+0.0075) & 0.7924(-0.0235) \\
    5 & 0.8367(-0.0121) & 0.7836(-0.0088) \\
    6 & 0.8326(-0.0041) & 0.8003(+0.0167) \\
    \bottomrule
  \end{tabular}
  \vspace{1em}
  \caption{Changes in variance}
  \begin{tabular}{crr}
    \toprule
    Agent & KCS & KCS+IBC \\
    \midrule
    1 & 0.1106(+0.0000) & 0.1120(+0.0000) \\
    2 & 0.1105(-0.0001) & 0.1126(+0.0006) \\
    3 & 0.1130(+0.0025) & 0.1138(+0.0012) \\
    4 & 0.1124(-0.0006) & 0.1168(+0.0030) \\
    5 & 0.1139(+0.0015) & 0.1176(+0.0008) \\
    6 & 0.1138(-0.0001) & 0.1160(-0.0016) \\
    \bottomrule
  \end{tabular}
\end{table}

To quantitatively evaluate the uncertainty of outputs and the diversity of opinions, we calculated the changes in the entropy and variance of the probability distributions at each agent step. Tables 4 and 5 show the changes in entropy and variance, respectively, for KCS and KCS+IBC. These values represent simple averages across the three datasets. This averaging aims to extract general trends attributable to model architecture, rather than characteristics specific to individual datasets.

In KCS, no clear consistency was observed in the step-by-step changes in entropy and variance. Instead, alternating increases and decreases were observed. For example, entropy increased for Agent 2, decreased for Agent 3, and then increased again for Agent 4, indicating that the direction of change repeatedly reversed. Similarly, the variance exhibited fluctuations at each step, and no clear trend of convergence or divergence was identified. These results suggest that sequential accumulation of information in KCS alone may be insufficient for aggregating judgments or resolving inconsistencies. This points to the potential need for supplementary mechanisms that enable reevaluation and reintegration of diverse perspectives during the opinion formation process.

In contrast, in KCS+IBC, entropy decreased consecutively by -0.0177 for Agent 3 and -0.0235 for Agent 4, with Step 3 showing the largest overall change. This step included a "chat session," which might have facilitated convergence of judgments through informal exchanges and reevaluation. Regarding variance, positive changes were observed throughout all steps except for the final analysis step (with a maximum increase of +0.0030), indicating a tendency toward increased dispersion. This trend stands in contrast to the decreasing entropy and suggests that while KCS+IBC enhances the confidence in judgments, it also maintains a structure that preserves minority opinions and diverse viewpoints. The results imply that the chat sessions in KCS+IBC enabled the kind of opinion recalibration and convergence that was difficult to achieve through sequential information transmission alone. The simultaneous assessment of entropy and variance confirms that KCS+IBC achieves a certain balance between convergence and diversity.

Figures 4 and 5, which show the transitions of means and standard errors, clearly illustrate differences in output dynamics between KCS and KCS+IBC.

\begin{figure}[h]
  \centering
  \includegraphics[width=1\columnwidth]{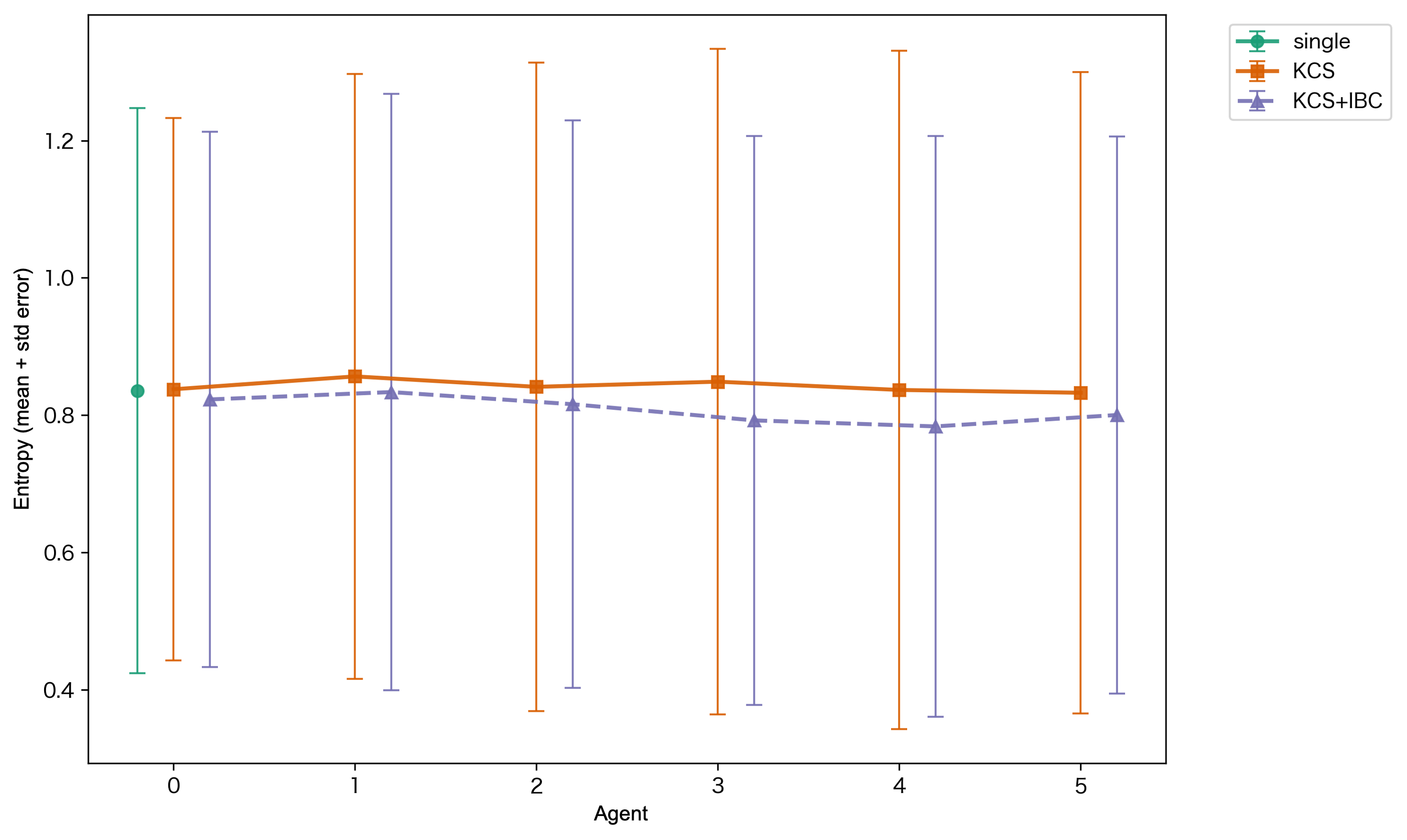}
  \caption{The mean and standard error of entropy for each agent}
  \label{fig:sample}
\end{figure}
\begin{figure}[h]
  \centering
  \includegraphics[width=1\columnwidth]{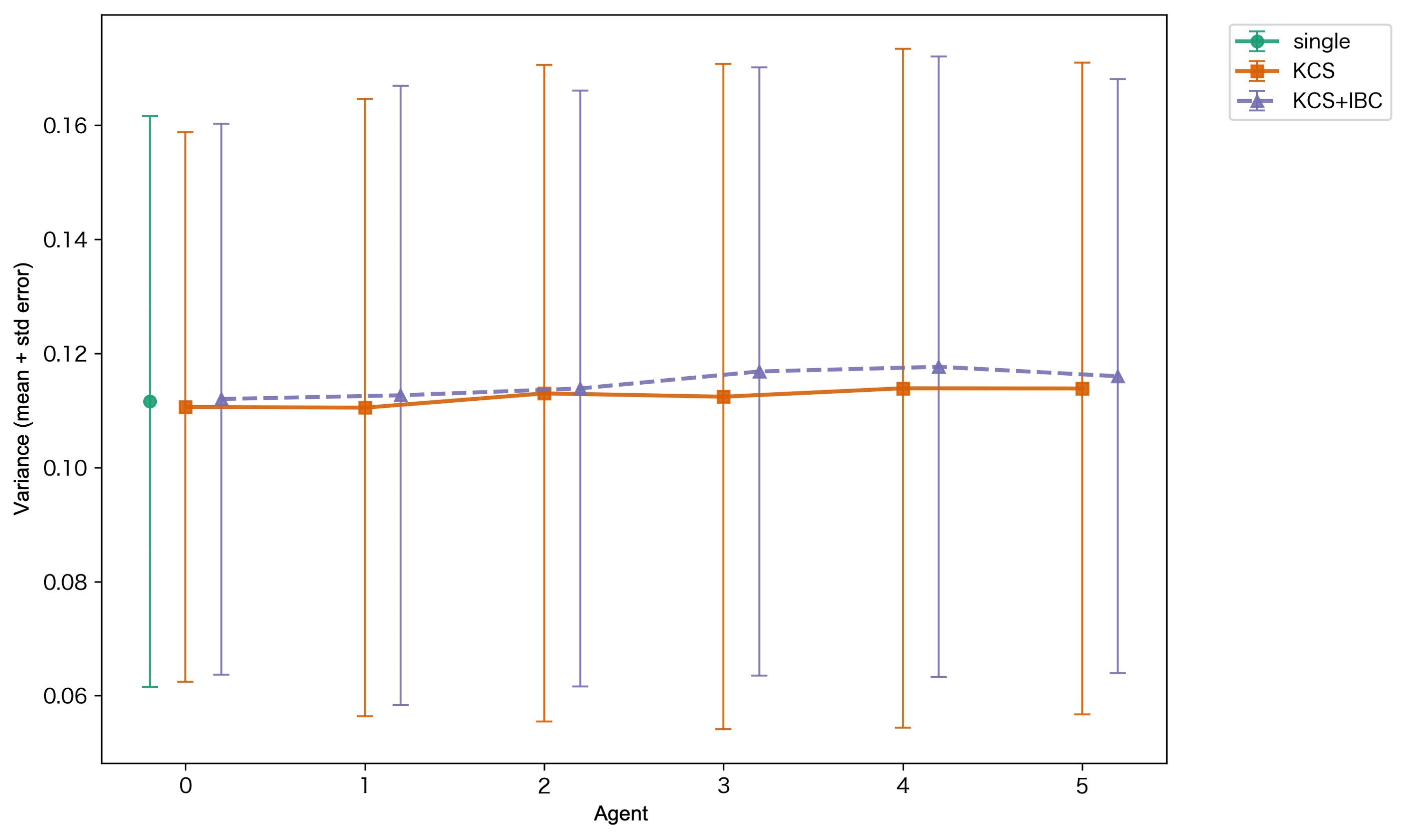}
  \caption{The mean and standard error of variance for each agent}
  \label{fig:sample}
\end{figure}
First, in KCS, both entropy and variance fluctuated irregularly across steps, with no discernible trend. This suggests that the process of accumulating opinions sequentially did not produce a consistent direction of convergence or divergence. Although information was being accumulated, the absence of a systematic mechanism for integration and reevaluation may have caused instability in the direction of judgments across agents.

On the other hand, in KCS+IBC, the mean entropy consistently decreased as the steps progressed, while the mean variance exhibited a gradual increasing trend. This pattern implies that the confidence in judgments increased, while the outputs retained a diversity of opinions. This tendency became more pronounced in the middle steps where chat sessions were introduced, indicating the impact of informal exchanges and reevaluation.

Another notable point is that KCS+IBC showed smaller and more stable standard errors compared to KCS, suggesting relatively higher consistency in outputs across agents. This indicates that KCS+IBC facilitates not only convergence of information but also a more stable opinion formation process.

Taken together, these findings lead to the conclusion that KCS+IBC enables outputs that combine increased confidence (through decreasing entropy), preservation of diversity (through increasing variance), and stability (through smaller errors) in multi-agent inference. This suggests that flexible reevaluation via informal dialogue mechanisms plays a key role in enhancing the quality of information aggregation, offering valuable insights for future multi-agent system design.

\section{Discussion}

This study proposed a multi-agent probabilistic inference framework inspired by a kairanban-style CoT system with idobata conversation and measured the entropy and variance of outputs for an emotion classification task. The results indicated a consistent decrease in entropy and a gradual increase in variance in the latter stages, suggesting that the framework possesses a structure capable of achieving both convergence of judgments and preservation of diversity. These findings demonstrate the effectiveness of a framework that simultaneously enhances confidence in judgments and maintains a range of perspectives. Specifically, decreasing entropy indicates improved inference confidence, while increasing variance suggests the retention of diverse viewpoints. This characteristic can be applied to decision-making support and model calibration, and visualizing the inference process can contribute to accountability.

On the other hand, the classification performance of the multi-agent probabilistic inference framework inspired by a kairanban-style CoT system with idobata conversation did not consistently outperform the single-agent baseline in ternary classification tasks such as the Financial Phrase Bank and TweetEval, with the exception of SST5. This suggests that collaborative inference does not always improve classification accuracy across all tasks, highlighting the need for careful tuning of prompt design and alignment with task characteristics when using LLMs. Additionally, structural parameters such as the placement of chat sessions, the number of agents, and the order of utterances were beyond the scope of this study and should be examined systematically in future work. Furthermore, it is important to test the applicability and effectiveness of this framework in tasks beyond emotion classification, such as summarization, opinion generation, and dialogue.

\section{Conclusion}

This study proposed a two-phase collaborative inference framework, KCS+IBC, inspired by the Japanese cultural practices of kairanban and idobata meetings, and demonstrated its ability to simultaneously control uncertainty and diversity in emotion analysis. The consistent decrease in output entropy during the later stages indicated enhanced confidence in judgments, while the gradual increase in variance suggested the retention of minority opinions and diverse perspectives. These findings support the conclusion that KCS+IBC provides a promising alternative to traditional sequential inference, contributing to reliability, explainability, and bias correction. Future work will involve quantitatively evaluating its calibration performance in probabilistic prediction and verifying its applicability across various tasks to further enhance its practical utility.

\bibliographystyle{ipsjsort}

\end{document}